\documentclass[letterpaper, 10 pt, journal, twoside]{IEEEtran}
\usepackage{amsmath,amsfonts}
\usepackage{algorithmic}
\usepackage{array}
\usepackage[caption=false,font=normalsize,labelfont=sf,textfont=sf]{subfig}
\usepackage{textcomp}
\usepackage{stfloats}
\usepackage{url}
\usepackage{verbatim}
\usepackage{graphicx}
\def\BibTeX{{\rm B\kern-.05em{\sc i\kern-.025em b}\kern-.08em
    T\kern-.1667em\lower.7ex\hbox{E}\kern-.125emX}}
\usepackage{balance}

\usepackage{microtype}
\usepackage{graphicx}
\usepackage{booktabs}
\usepackage[utf8]{inputenc}
\usepackage{hyperref}

\usepackage{tabularx}
\usepackage{algorithm}
\usepackage{algorithmic}
\usepackage{amsmath}
\usepackage{amssymb}
\usepackage{mathtools}
\usepackage{amsthm}
\usepackage{multirow}
\usepackage{multicol}
\usepackage{overpic}
\usepackage[capitalize,noabbrev]{cleveref}
\usepackage{caption} 

\theoremstyle{plain}

\theoremstyle{definition}

\theoremstyle{remark}

\usepackage[textsize=tiny]{todonotes}
\usepackage{cuted} 
\usepackage{hyperref}

\usepackage{pifont}

\usepackage{enumitem}
\setlist[enumerate]{leftmargin=*}
\setlist[itemize]{leftmargin=*}

\newcommand{\simtoreal}{sim-to-real}

\newcommand{\name}{SLIM}

\newcommand{\Search}{{\footnotesize\ttfamily Search}}
\newcommand{\MoveTo}{{\footnotesize\ttfamily MoveTo}}

\newcommand{\Grasp}{{\footnotesize\ttfamily Grasp}}
\newcommand{\SearchWithObj}{{\footnotesize\ttfamily SearchWithObj}}
\newcommand{\MoveToWithObj}{{\footnotesize\ttfamily MoveToWithObj}}
\newcommand{\MoveGripperToWithObj}{{\footnotesize\ttfamily MoveGripperToWithObj}}
\newcommand{\DropInto}{{\footnotesize\ttfamily DropInto}}

\newcommand{\NoArmRetract}{{\textbf{No Arm Retract}}}
\newcommand{\NoPerturb}{{\textbf{No Perturbation}}}
\newcommand{\NoVisualAug}{{\textbf{No Visual Aug}}}
\newcommand{\DistillationOnly}{{\textbf{Distillation Only}}}
\newcommand{\HumanTeleop}{{\textbf{Human Teleop}}}
\newcommand{\NoDistillation}{{\textbf{No Distillation}}}

\newcommand{\bkcolor}[1]{\mathbf{#1}}

\newcommand{\tikzxmark}{%
\tikz[scale=0.18] {
    \draw[line width=0.7,line cap=round] (0,0) to [bend left=6] (1,1);
    \draw[line width=0.7,line cap=round] (0.2,0.95) to [bend right=3] (0.8,0.05);
}}
\newcommand{\tikzcmark}{%
\tikz[scale=0.18] {
    \draw[line width=0.7,line cap=round] (0.25,0) to [bend left=10] (1,1);
    \draw[line width=0.8,line cap=round] (0,0.35) to [bend right=1] (0.23,0);
}}

\newcommand{\piagent}{\pi_{\rm stu}}
\newcommand{\piexpert}{\pi_{\rm tea}}

\newcommand{\Figure}{Fig.}

\usepackage{dcolumn}
\newcolumntype{e}[1]{D{+}{\,\pm\,}{#1}} 
\newcolumntype{Y}{>{\centering\arraybackslash}X}

\newcommand{\secref}[1]{\textbf{\S{#1}}}

\begin{document}

\title{Learning Multi-Stage Pick-and-Place with a \\ Legged Mobile Manipulator}

\markboth{IEEE Robotics and Automation Letters. Preprint Version. Accepted August, 2025}
{Zhang \MakeLowercase{\textit{et al.}}: Learning Multi-Stage Pick-and-Place with a Legged Mobile Manipulator}

\author{Haichao Zhang, Haonan Yu, Le Zhao, Andrew Choi, Qinxun Bai, Yiqing Yang,  Wei Xu

\thanks{Horizon Robotics
        {\tt\footnotesize \{first\_name.last\_name\}@horizon.auto}}%
}

\IEEEaftertitletext{%
  \noindent\begin{minipage}{\textwidth}\centering
    \includegraphics[width=\textwidth]{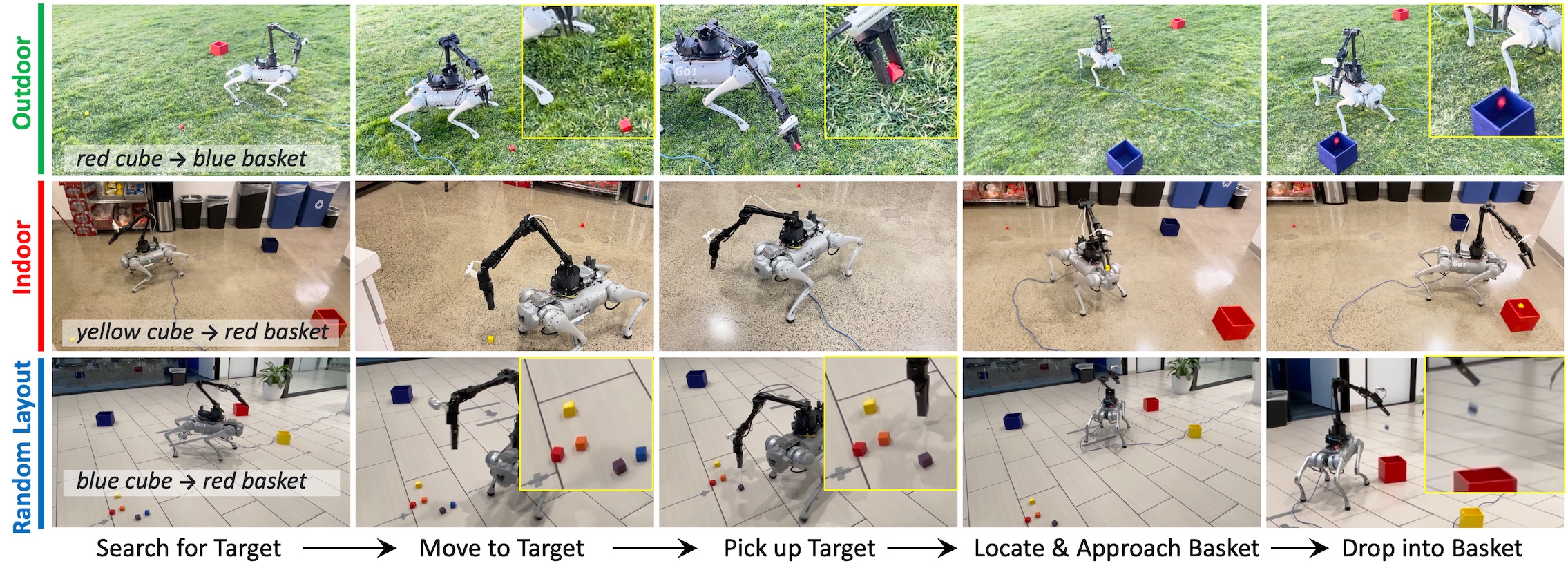}
    \captionof{figure}{\textbf{Multi-Stage Pick-and-Place in Real.} Snapshots of the \emph{same} visuo-motor policy deployed in diverse real-world scenes, featuring significant variations in terrain, background, distractors, and other environmental factors.
        The mobile pick-and-place task involves multiple stages as annotated at the bottom of the figure. Note that these annotations are added for understanding task progress and are not part of the input to the visuo-motor policy.
        More demo videos are available on the \href{https://horizonrobotics.github.io/gail/SLIM}{project page}.
    }
    \label{fig:multi_stage_pick_and_place}
  \end{minipage}\par\medskip
}

\maketitle

\begin{abstract}
Quadruped-based mobile manipulation presents significant challenges in robotics due to the diversity of required skills, the extended task horizon, and partial observability. 
After presenting a multi-stage pick-and-place task as a succinct
yet sufficiently rich setup that captures key desiderata for
quadruped-based mobile manipulation, we propose an approach that can
train a visuo-motor policy entirely in simulation, and achieve nearly 80\% success in the real world. The policy efficiently performs search, approach, grasp, transport, and drop into actions, with emerged behaviors such as re-grasping and task chaining.
We conduct an extensive set of real-world experiments with ablation studies highlighting key techniques for efficient training and effective sim-to-real transfer.  Additional experiments demonstrate deployment across a variety of indoor and outdoor environments. 
Demo videos and additional resources are available on the project page: \url{{https://horizonrobotics.github.io/gail/SLIM}}.
\end{abstract}

\begin{IEEEkeywords}
Mobile manipulation, Reinforcement Learning, Legged Robots, Deep Learning in Grasping and Manipulation.
\end{IEEEkeywords}

\section{Introduction}
\IEEEPARstart{Q}{uadruped-based} mobile manipulation holds great potential by combining the ability of traversing diverse scenes using a legged base with the manipulation ability of an arm mounted on the base. It enables robots to perform a wide variety of complex tasks in diverse environments~\cite{deep_wholebody,ASC, pan2024roboduet,VBC,GEFF, GAMMA}. 
Effective learning of visuo-motor policies for mobile manipulation is the key to unlocking these capabilities.
Recent attempts utilize sim-to-real for quadruped mobile manipulation~\cite{deep_wholebody, pan2024roboduet, VBC, GAMMA}. However, they focus either on improving the task-agnostic low-level capabilities~\cite{deep_wholebody, pan2024roboduet}, or on solving \emph{short-horizon} mobile manipulation (\emph{e.g.} move to an object in view and pick it up)~\cite{VBC, GAMMA}.
How to generalize to handle longer-horizon tasks remains a significant challenge.

We present \name{} (\textbf{S}im-to-Real \textbf{L}earning of  Long-Hor\textbf{i}zon  Legged \textbf{M}anipulation), a complete system for training robotic policies entirely in simulation and deploying them zero-shot in the real world. 
We benchmark the proposed approach on a multi-stage mobile pick-and-place task consisting of multiple stages: search, move to, grasp, transport, and drop into, achieving nearly 80\% real-world success.
We uncover a number of key ingredients for achieving successful real-world long-horizon task completion.
\noindent Our key contributions are:
\begin{enumerate}[itemsep=0.1em]
    \item We present an approach for learning to solve the multi-stage pick-and-place task for a legged mobile manipulator; 
    \item We identify the challenges of long-horizon task learning and introduce progressive policy expansion for long-horizon task learning;
    \item We conduct an extensive set of real-world experiments with 400 episodes across various indoor / outdoor scenes;
    \item Our pipeline trains models fully in simulation, and achieves $\sim$80\% success in the real world, with great generalization across diverse scenes.
\end{enumerate}

\vspace{-0.03in}
\section{Related Work}\label{sec:related_work}
\vspace{-0.02in}

Legged mobile manipulation has been an active research topic in the robotics community~\cite{review1, review2}, driven by the potential of legged systems to traverse unstructured environments while interacting with objects.
Classical approaches typically design wholebody controllers or combine a locomotion policy with model-based manipulation control~\cite{sleiman2023wholebody_control, ma2022mpc}.

Recent advances in reinforcement learning (RL) based legged locomotion have prompted researchers to push the boundaries of learning-based legged mobile manipulation. 
One line of research focuses on training low-level policies to replace hand-designed loco-manipulation controllers~\cite{deep_wholebody, pan2024roboduet, huang2024manipulatortail}, with quick demonstrations of their applicability to high-level tasks via teleoperation. The potential issues and challenges of using such
policies for high-level task learning fall outside the scope of
these works and remain unclear.

Another area of research trains task-aware policies for mobile manipulation, showing encouraging results.  However, learning and maintaining high success throughout full task execution on top of a noisy low-level policy presents a great challenge.  Existing research typically focuses on short-horizon tasks with only a small number of stages~\cite{wholebody, zhang2024learning, VBC}, so that they are within the reach of
standard RL algorithms. 

While making encouraging progress, the challenge of solving long horizon tasks with multiple stages—common and often unavoidable in real-world scenarios—remains largely unaddressed.
Some other work simplify the task by removing partial observability, either add a global camera to observe the whole scene~\cite{wu2024helpfuldoggybot}, or require that the target object be visible the whole time~\cite{VBC}. Such assumptions restrict the robot to environments with specific camera setups or require careful staging, thus limiting the applicability of the approach compared to methods that rely solely on onboard sensors and do not assume initial object visibility.

ASC \cite{ASC} also targets multi-stage, long-horizon tasks, with a focus on learning to chain skills such as navigation, picking and placing, many of which are supported by external APIs (\emph{e.g.} Boston Dynamics API). As a result, task complexity and learning difficulties are greatly reduced.
Also, techniques used there (\emph{e.g.} floating-base approximation to the low-level policy) have a strong dependency on the assumed behavior of the external APIs, limiting its usage in more general cases. 
ASC further reduces the difficulty of the task by including in the observation \textbf{(1) robot’s relative heading and displacement from
its start pose} (\emph{i.e.} robot's global position wrt a global coordinate); \textbf{(2) the target position} (the approximate receptacle location where the object can be picked or placed).
These special requirements largely makes the observation globally observable, reduces the difficulties of the task and makes the system confined to lab-like  environment and less general to be deployed in a general environment where no special sensors are deployed (\emph{e.g.} outdoor).

Different from prior work, we focus on learning a complete visuo-motor policy fully in simulation to solve multi-stage, long-horizon tasks with a quadruped mobile manipulator using only onboard sensors, and explicitly address the complexities and challenges of long horizon task learning under this setting.

\begin{figure}[h]
    \centering
    \begin{overpic}[width=8.6cm]{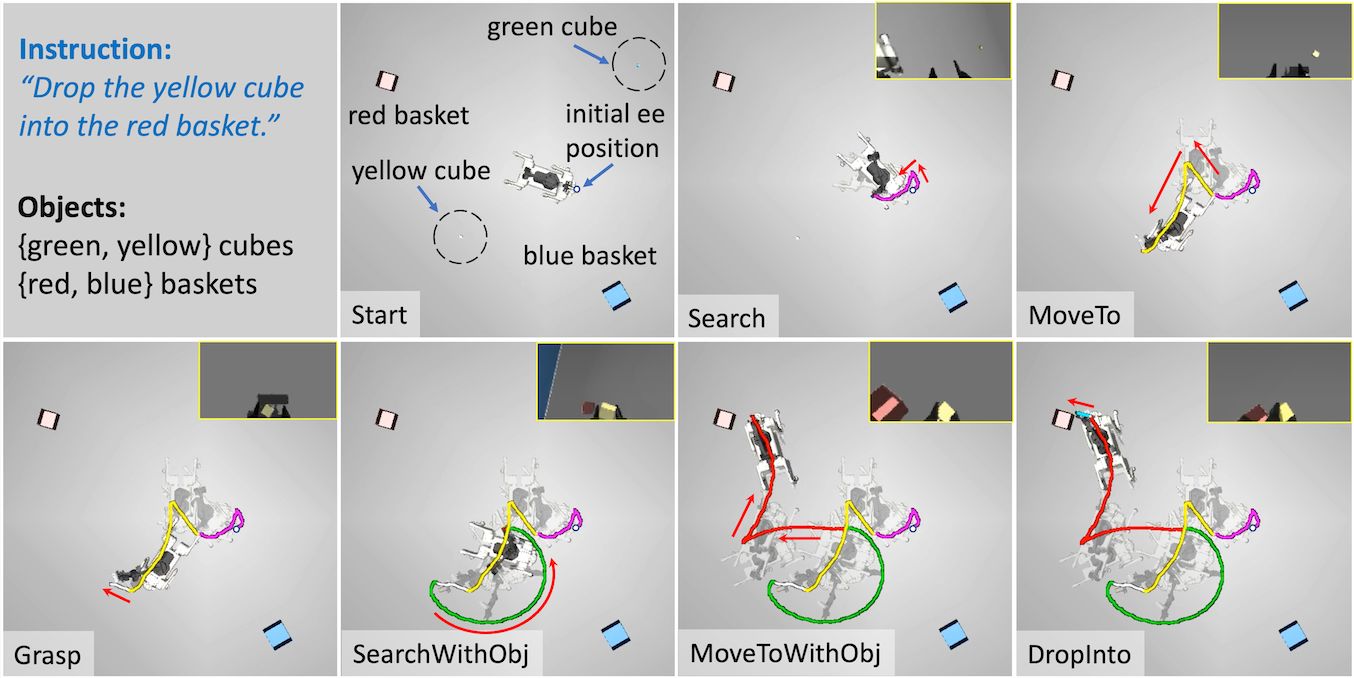}
    \end{overpic}
    \caption{\textbf{Multi-Stage Pick-and-Place Task Visualization.} 
    Example rollout in simulation of a teacher policy (Section~\ref{sec:teacher_learning}). For visualization purpose, we removed background objects and textures from the scene. We also visualize the EE trajectories. Trajectories corresponding to different stages are rendered with different colors. The ego-centric camera view is shown on the top-right of each image. The red colored arrows denote the direction of movement.}
    \label{fig:task_visualization}
    \vspace{-0.15in}
\end{figure}

\vspace{-0.04in}
\section{The Multi-Stage Mobile Pick-and-Place Task}
Quadruped-based mobile manipulation encompasses a broad range of potential tasks. However, to the best of our knowledge, there is currently no standardized benchmark protocol specifically designed to evaluate such systems. Given our focus, we design a custom task to validate our method, guided by the following desiderata that are important towards learning to solving real-world mobile manipulation tasks:
\begin{itemize}[label={}, leftmargin=1.em, align=left]
    \item[\secref{1}] {\bf long-horizon:} the task should consist of multiple stages, as commonly encountered in real-world applications, testing the long-horizon learning ability; 
    \item[\secref{2}] {\bf ego-centric:} it should rely solely on onboard sensors, resulting in partial observability;
    \item[\secref{3}] {\bf active search:} it should not assume that the target object is visible at the beginning of the episode, requiring the ability to actively search for the target;
    \item[\secref{4}] {\bf practical diversity:} it should support a reasonable amount of scene variations for meaningful generalization testing while maintaining a minimal complexity, making it easy to set up and repeat across different environments. 
\end{itemize}

\begin{figure}[t]
    \centering
    \vspace{0.08in}
    \begin{overpic}[width=8.5cm]{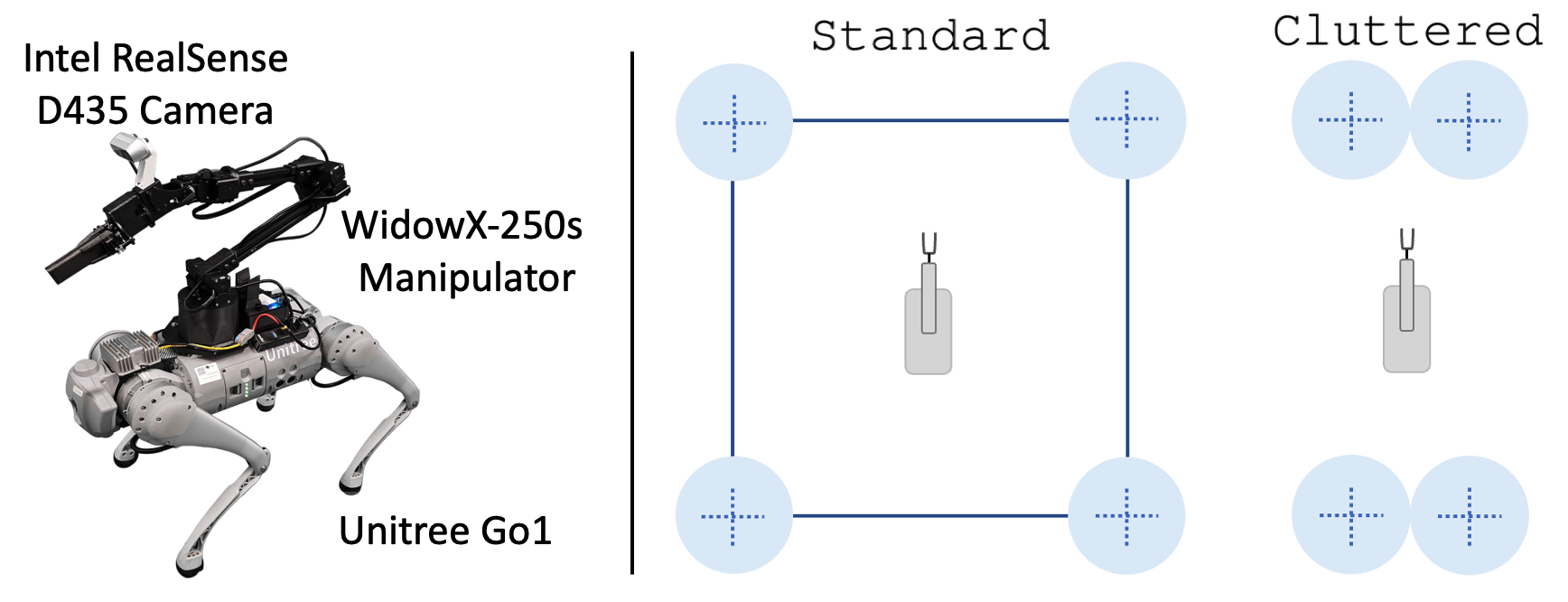}
    \end{overpic}
    \vspace{-0.1in}
    \caption{\textbf{Robot and Scene Spatial Layouts for Multi-Stage Pick-and-Place.} 
    {\footnotesize{\ttfamily Standard}}: we put the robot in the center of a $2{\rm m}\! \times\! 2{\rm m}$ square space and objects including objects to be grasped and containers to drop the graspable objects into on the four corners. 
    During training, the spatial layout used for the objects are randomize based on the {\footnotesize{\ttfamily Standard}} setting, with each position sampled from a circular region centered around each corner with a radius of $0.5$m.
    {\footnotesize{\ttfamily Cluttered}}: objects are placed close to each other in front and behind the robot, leading to a spatial layout that is out of the training distribution (OOD). The instructions and object placements vary across episodes for both the {\footnotesize{\ttfamily Standard}} and the {\footnotesize{\ttfamily Cluttered}} settings. 
    }
    \vspace{-0.2in}
    \label{fig:robot_and_scene_config}
\end{figure}

\noindent Given these desiderata and inspired by the important role of \emph{pick-and-place} task in stationary tabletop manipulation~\cite{tabletop_pick_place}, we propose a \emph{multi-stage mobile pick-and-place} task (\Figure~\ref{fig:task_visualization}).
In this task, the robot, the target objects together with distractors are all placed in an area on the floor. 
The robot needs to accomplish multiple stages to solve the task (\secref{1}):   first search for the target object on the ground, go to the object, pick it up, then search for the target container while holding the object in gripper, carry the grasped object to the container, and drop the object into the container. 
Everything is done using only one wrist-mounted camera (\Figure~\ref{fig:robot_and_scene_config}), which is ego-centric and creates partial observability (\secref{2}).
The objects are placed around the robot (\Figure~\ref{fig:robot_and_scene_config}) and are not necessarily visible initially (\secref{3}). Additionally, after pick up, the target container is not placed in a fixed location, requiring search again.
To introduce scene variations, we introduce distractors into the scene,
and specify the target object and container with a language instruction at the beginning of the task (\secref{4}). To maintain a minimum level of complexity, we use colored cubes as the objects to be grasped and colored baskets as the container following~\cite{tabletop_pick_place}. Instructions follow this format: ``{\footnotesize \texttt{Drop the \{color1\} cube into the \{color2\} basket}}", where {\footnotesize \texttt{\{color1\}}} and {\footnotesize \texttt{\{color2\}}} are sampled from {\footnotesize \texttt{\{red, blue, green, yellow\}}}.

Given desiderata \secref{1}$\sim$\secref{4} and their corresponding implementations in the proposed multi-stage pick-and-place task, not only can we quantitatively measure and compare policy performance under repeatable variations (Table~\ref{tab:real_world_experimental_results}), but also test the generalization ability across diverse real-world scenes (\Figure~\ref{fig:multi_stage_pick_and_place}), featuring large variations in terrain, background, distractors, and other environmental factors.

\begin{figure*}[t]
    \centering
    \vspace{0.08in}
    \begin{overpic}[width=\textwidth]{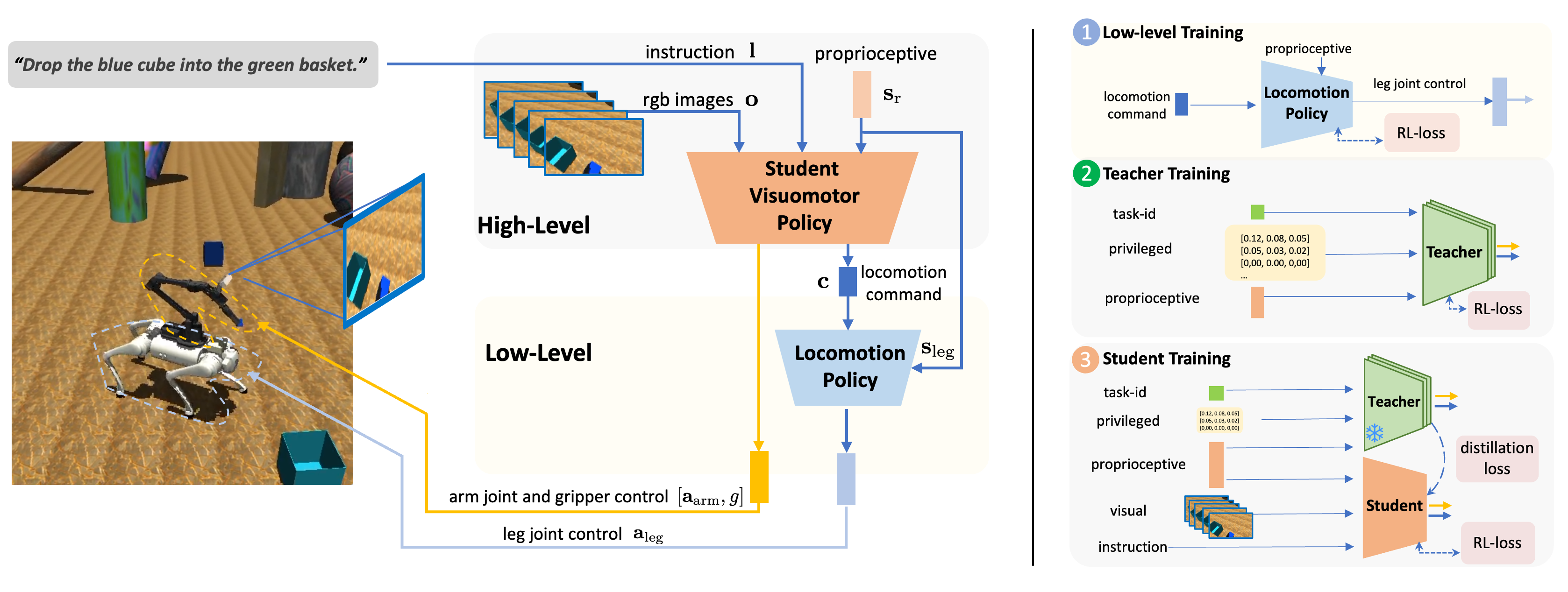}
    \end{overpic}
    \vspace{-0.35in}
    \caption{\textbf{Hierarchical Framework and Visuomotor Policy Pipeline (Left).} Given language instruction and sensor inputs, the high-level policy generates a set of two control signals: 1) the arm and gripper control signals, and 2) the locomotion command. The arm control signals are directly passed to the arm driver, and the locomotion command is passed to the low-level policy to control the leg joints of the quadruped. 
    \textbf{Low-Level Training (Top Right), and High-Level Teacher and Student Training (Middle and Bottom Right).} \name{} training is divided into three sequential stages. First, the low-level locomotion policy is trained via RL to follow a sampled linear and angular velocity command. 
    Second, the high-level teacher policy is trained via RL with privileged low-dimensional state input to solve the long-horizon task. 
    Finally, the student visuomotor policy is trained by distilling the teacher behavior while maximizing task rewards, using visual, sensory, and language instruction as input. 
    Both the teacher and student policies command the same frozen low-level policy produced by the first stage.  The teacher is only run in simulation and the student can be deployed in real.
    }
    \label{fig:framework}
    \vspace{-0.15in}
\end{figure*}

 \vspace{-0.1in}
\section{Method}
In order to successfully operate the robot hardware for both fast locomotion control and visual task level reasoning, the autonomous policy is divided into two levels: high-level visuomotor control and low-level locomotion control.
The low-level policy takes over the responsibilities of legged locomotion, trained following \cite{walk_these_ways} and frozen in high-level training, while the high-level focuses on task-dependent decision making.  The high-level policy needs to process high dimensional visual and language input, and complete the long-horizon task by sending locomotion and arm commands to the low-level (\Figure~\ref{fig:framework} left).  The high-level still presents an enormous space for the policy to explore, and a challenge for efficient RL training.  We further adopt the widely used teacher-student framework \cite{zhang2024learning, fan2021secant} for high-level learning (\Figure~\ref{fig:framework} middle and bottom right).
Directly training a visuomotor policy using RL is very challenging, due to the compounded challenges of RL and representation learning from raw pixel observations. By leveraging the teacher-student framework, we decouple the challenges by first learning to solve the RL task without the additional difficulties from the representation learning (teacher learning stage);  and then learn the visuomotor policy in the student learning stage, guided by the teacher, largely avoiding the difficulty of RL exploration when learning with image observations.

The 
teacher is trained purely with RL from privileged, structured, and low-dimensional inputs.
The privileged observation includes the object's positions and orientations. Note that visibility mask is applied: object features that are outside of the camera view are set to zero and only the features of the objects that are visible in the field of camera view are retained  (\Figure 4 Teacher Training).
When a teacher policy is successfully learned, it is frozen and used to guide the student via behavior distillation.
As the student will eventually be deployed in the real world, it does not rely on any privileged information.
In the two sections below, we explain the teacher and the student policies in more detail.

\subsection{The Teacher: Long-Horizon Task Learning}
\label{sec:teacher_learning}

The Teacher works with privileged object state information to solve the full task.  By offloading the complexities of visual representation learning to the student, it mainly focuses on the following challenges of long-horizon task learning,
\begin{enumerate}
    \item \textbf{\textit{Loss of capacity and catastrophic forgetting}}: 
    As the teacher progresses in training, it needs to cope with both the loss of capacity~\cite{capacity_loss, understanding_plasticity}, which hinders the network from continual learning,
    and the catastrophic forgetting issue, which could destroy skills that have already been acquired.
    \item \textbf{\textit{Continual exploration}}: 
    For long-horizon tasks, there are usually a number of milestones that must be sequentially achieved.
    Therefore, even after reaching one milestone, the teacher must continue to explore in order to solve the next stage.
    Without carefully encouraging continual exploration, the teacher can stop exploration early and settle on a suboptimal solution.
\end{enumerate}
Addressing these challenges is the key for successful policy learning for long-horizon tasks.  We propose to use \emph{task decomposition} and further integrate with \emph{policy expansion}~\cite{pex} for this purpose.

{\noindent \emph{A.1} \bf{Task Decomposition.}}
The long-horizon task $\mathcal{T}$ can be decomposed into $K$ subtasks $\{\tau^k\}_{k=1}^K$ with shorter horizons.
Using the multi-stage pick-and-place task as an illustration, an example decomposition can be {\{\Search{}, \MoveTo{}, \Grasp{}, \SearchWithObj{}, \MoveToWithObj{}, \MoveGripperToWithObj{},  \DropInto{}\}}.
Task decomposition offers several benefits: 1) a modular design of task rewards. Apart from the common rewards such as sparse subtask success reward and distance-based shaping reward to encourage exploration toward the target, additional shaping rewards can also be easily incorporated for each task. For example, for \SearchWithObj{} and \MoveToWithObj{}, we add an arm-retract reward (another distance-based shaping reward), to encourage the arm to stay close to a canonical pose shown in \Figure~\ref{fig:multi_stage_pick_and_place}. 
This reward serves to promote good vantage for the wrist camera and discourages other suboptimal behaviors after grasping such as looking at the ground during the \SearchWithObj{} subtask.
Its impact is investigated in the ablations in Section~\ref{sec:result} (\NoArmRetract{} baseline).
2) task decomposition also allows various behavior priors to be conveniently integrated into the policy. For example, we use two types of behavior priors in this work: stationary manipulation and rotational search, enabling the agent to remain stationary during manipulation, or avoid walking off the workspace during search,  which are key for both safe and successful deployments (Section~\ref{sec:main_results}).
3) it can be integrated with a properly learning scheme to address the long-horizon learning challenge, as shown below.

{\noindent \emph{A.2} \bf{Progressive Policy Expansion.}}
To incorporate the task decompositional structure in learning, we leverage the fact that privileged information 
is accessible to the teacher and thus include the subtask index $k$ as part of the privileged observation.
Note the final visuo-motor policy (Seciton~\ref{sec:student}) does not use subtask id as input.

\begin{figure}[t]
    \centering
    \includegraphics[width=\linewidth]{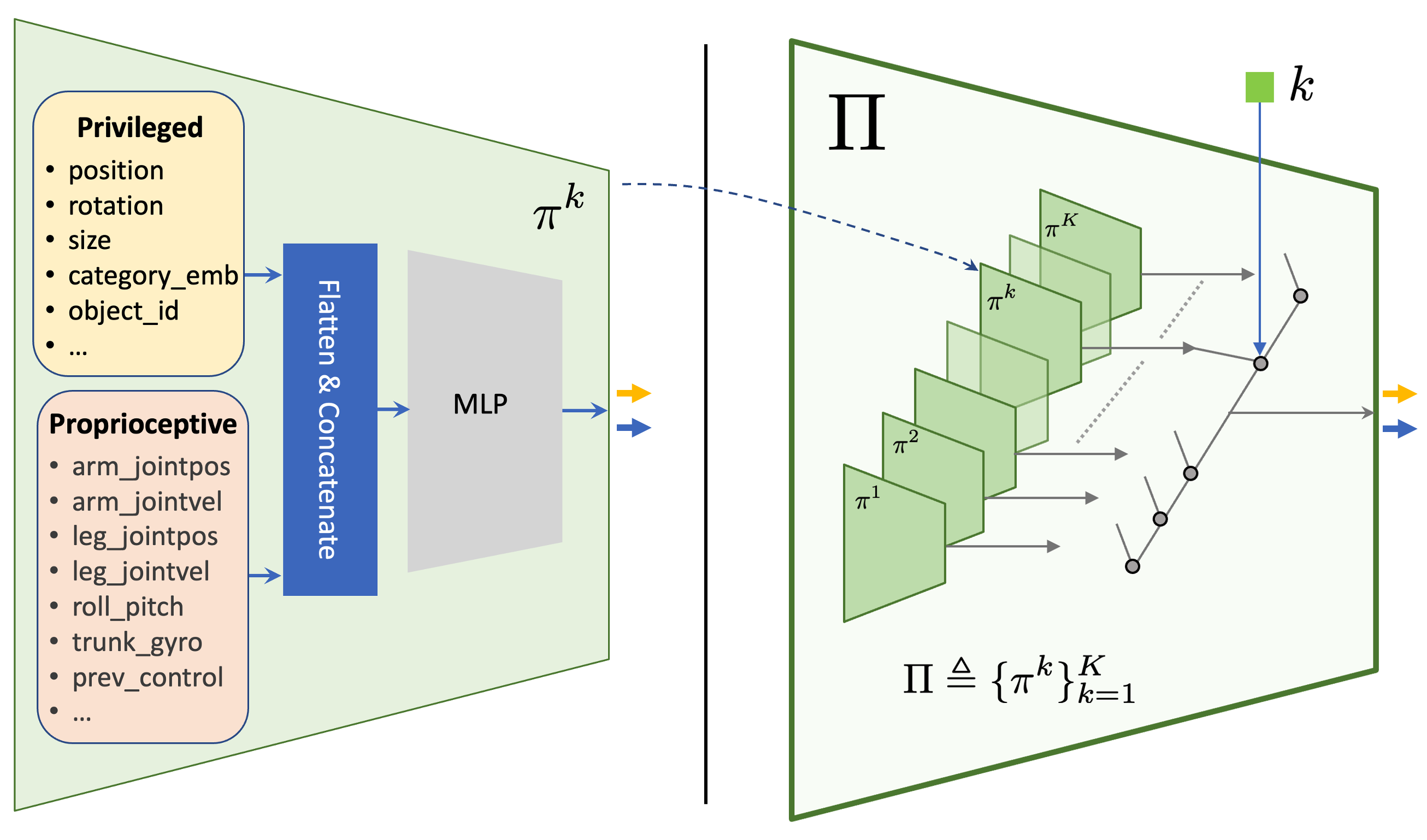}
    \vspace{-0.2in}
    \caption{\textbf{Teacher Policy Network Structure.}
    The full teacher network $\Pi$ is a set of structurally identical networks $\{\pi^{k}\}_{k=1}^K$ gated by the subtask id $k$.
    On the left, each individual teacher policy $\pi^k$ takes a set of privileged and proprioceptive input, flattening and concatenating them before passing it through an MLP. 
    On the right, we show the full teacher network $\Pi$. Given an subtask id $k$, the $k$-th individual teacher policy is activated during computation, \emph{i.e.,} $\Pi[k] \equiv \pi^{k}$.
    }
    \vspace{-0.2in}
    \label{fig:expert_nn}
\end{figure}
We can solve the long-horizon task by activating the network instance among a set of policies  according to the task id. 
This way, there will be dedicated policies for continual exploration and learning of new subtask without affecting any skills acquired in the previous subtasks.
Intuitively, it works as follows. 
We initiate the exploration and learning with a single active policy network $\Pi=\{ \pi^1\}$ that is responsible for learning to solve the initial subtask.
Whenever a new subtask is encountered a new policy is added into the active policy set, \emph{i.e.,} $\Pi=\{\pi^1, \pi^2\}$.
By doing this progressively for all the $K$ subtasks, we get  
\begin{align}\label{eq:expert}
    \Pi &\triangleq \{ \pi^k\}_{k=1}^K,
\end{align}
where $\pi^k$ denotes an individual teacher policy for subtask $k$, and $\Pi$ denotes the full teacher network comprised of a set of individual teacher policies,
as illustrated in \Figure~\ref{fig:expert_nn}.
This can be regarded as a progressive application of policy expansion (PEX)~\cite{pex}, and we dub it Progressive PEX.

Policy expansion at the initiation of each subtask naturally addresses the two challenges brought by long-horizon task learning:
1) it can achieve multi-stage continual exploration, and the full ability of exploration is guaranteed when entering a new stage; 2) since the policy for solving one subtask is encapsulated in a dedicated network, it mitigates catastrophic forgetting of the already learned skills. For a newly encountered subtask, the dedicated network that will be newly allocated addresses the issue of lost plasticity.

The left side of \Figure~\ref{fig:expert_nn} provides an illustration of one individual teacher policy network $\pi^k$, taking low-dimensional proprioceptive and privileged observation as input and is responsible for learning to solve the corresponding subtask.
The right side of \Figure~\ref{fig:expert_nn} shows the full teacher network 
$\Pi$.  A forward pass through $\Pi$ is carried out by the corresponding sub-network in $\Pi$ indexed by the subtask id $k$: $\Pi[k] \equiv \pi^{k}$.

\begin{figure}[t]
    \centering
    \includegraphics[width=1\columnwidth]{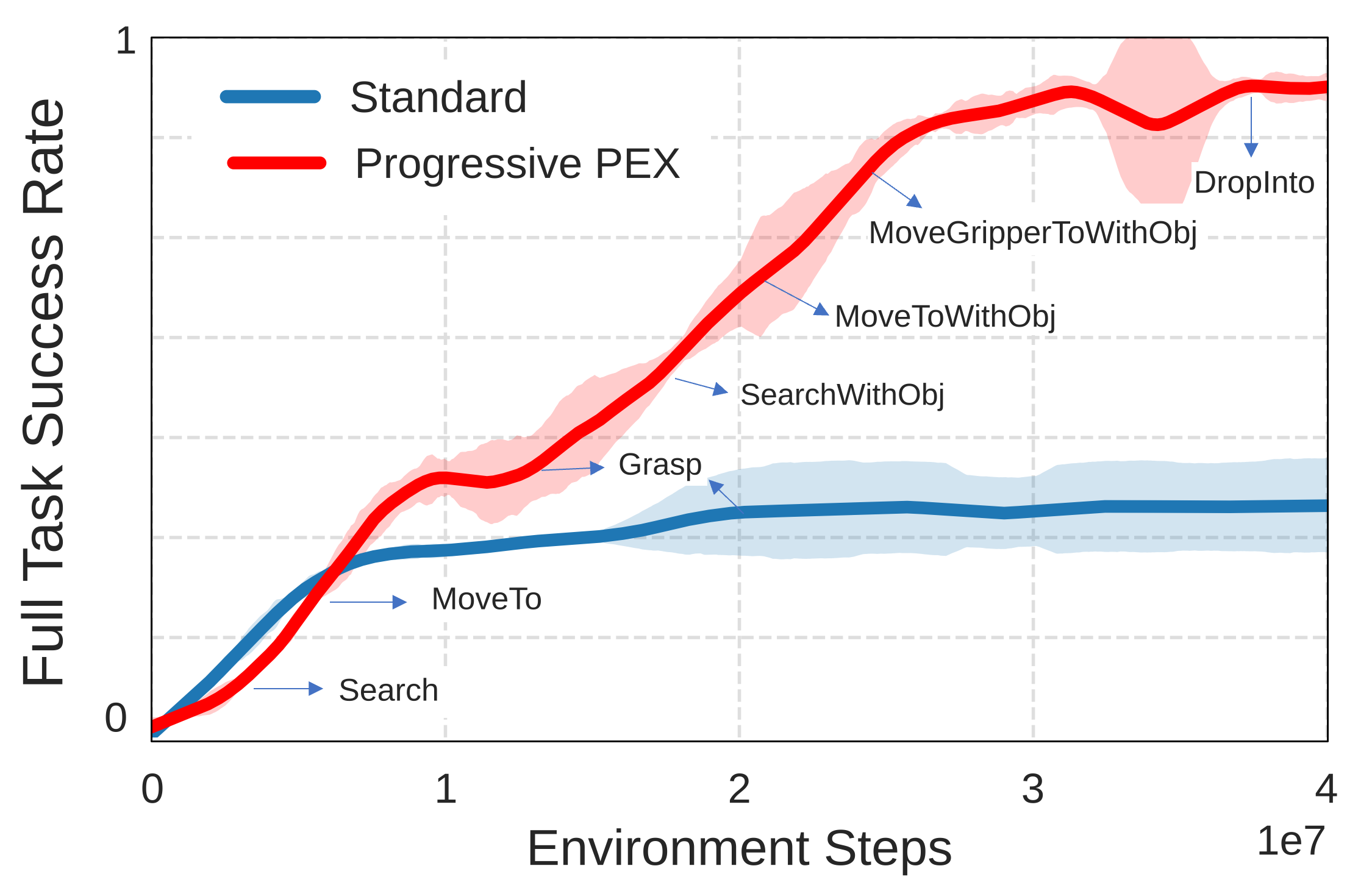}
     \vspace{-0.2in}
    \caption{\textbf{Standard Method v.s. Progressive PEX}. We compare the learning behaviors of the Standard method (without progressive policy expansion) and the Progressive PEX approach in terms of the full task success rate as training proceeds (the mean and standard deviation calculated across 3 seeds).
    We also provide rough annotations of the subtask stages that are under learning along the curve.
    }
    \label{fig:pex_vs_standard}
    \vspace{-0.25in}
\end{figure}

The training of $\Pi$ is done via a multi-task variant of SAC~\cite{sac, metaworld, PaCo}.  
The performance comparison between the method using a standard network structure and the Progressive PEX approach on solving long-horizon tasks is provided in \Figure~\ref{fig:pex_vs_standard}, which clearly shows Progressive PEX has a much stronger ability in long-horizon task learning.

\vspace{-0.1in}
\subsection{The Student: Policy Distillation Guided RL}
\vspace{-0.05in}
\label{sec:student}

The student policy is the actual high-level policy that gets deployed in the real world.
It perceives the surrounding environment using the single RGB stream and motor sensors, and follows language instructions.

We train the student by modifying SAC~\cite{sac} to incorporate the distillation loss properly.
First, we use a mixed rollout strategy to generate replay data. 
At the beginning of a new episode, with a certain probability we will sample actions for the entire episode from the student policy, or from the teacher policy.
On the one hand, high-performing trajectories from the teacher along the whole task horizon facilitates coverage of future subtasks that the student cannot yet solve.
On the other hand, student rollout allows the student to explore and cover more state space.

Second, we remove the entropy reward from policy evaluation~\cite{yu2022you} and policy improvement, replacing it with the distillation loss with a fixed weight $\alpha$:
\begin{equation}
\begin{array}{l}
\displaystyle\max_{\piagent}\displaystyle\mathbb{E}_{(\mathbf s_\textrm{stu},\mathbf{s}_\textrm{p},k)\sim \mathcal{D}_{\text{replay}}}
\Big[\mathbb{E}_{\mathbf{a}_{\text{hi}}\sim\piagent(\cdot|\mathbf s_\textrm{stu})}Q\big(\mathbf s_\textrm{stu},\mathbf{a}_{\text{hi}}\big)\\
\ \ \ \ \ \ \ -\alpha {\rm KL}\big(\piexpert(\mathbf{a}_{\text{hi}}|\mathbf{s}_\textrm{r},\mathbf{s}_\textrm{p}) \, || \, \piagent(\mathbf{a}_{\text{hi}}| \mathbf s_\textrm{stu}) \big)\Big],\\
\end{array}
\label{eq:student}
\end{equation}
where $\piexpert\!\triangleq\!\Pi$ and $\piagent$ denote the teacher and student policies respectively,
$\mathbf s_\textrm{r}$ is the robot's proprioceptive state and $\mathbf s_\textrm{p}$ the privileged observation (\textit{e.g.}, object positions, categories, subtask id \textit{etc.}).
$\mathbf s_\textrm{stu}\!=\! [\mathbf o, \mathbf s_\textrm{r}, \mathbf l]$, with $\mathbf o$, $\mathbf s_\textrm{r}$ and $\mathbf l$ denoting the 
(stacked) ego-centric RGB images, proprioceptive state vectors, and the language instruction.
$\mathbf{a}_{\text{hi}}$ is the high-level action including the locomotion command (representing the target forward and angular velocities for the quadruped), the arm delta joint position and the gripper control signal.

To ensure that the KL term encourages the student policy $\piagent$ to explore, we modify the teacher policy $\piexpert$, keeping the mode of the action distribution unchanged but assign a fixed modal dispersion $\sigma$ (\emph{e.g.,} std. for Gaussian).
This results in a distillation loss that achieves two goals: imitating the mode while encouraging exploration.

\vspace{-0.15in}
\subsection{Sim-to-Real Gap Reduction}
Policies trained in simulation can fail in real due to the notorious sim-to-real gap.  We identify key techniques for successful real deployment.

{\noindent \emph{C.1} \bf{Dynamics Gap.}} The dynamics \simtoreal{} gap is caused by the mismatch in the physical properties of objects and motors, or by things that are not properly simulated, such as friction and backlash.
This difference in dynamics can cause policies trained purely in simulation to fail when deployed in the real world, especially if the task requires accurate motor control, such as grasping a small object as in the multi-stage pick-and-place task considered in this work.
For reducing dynamics \simtoreal{} gap, we used 1) PID arm control to minimize tracking error, 2) arm control perturbation by adding random noise to the arm joint position targets, 3) arm mount perturbation by randomly perturbing the arm mount position and yaw at the beginning of each episode to make the learned policy robust to \simtoreal{} gaps that may arise from the actual mounting accuracy, torso height variations, \emph{etc}, 4) object perturbations, to enforce the learning of reactive policies robust to object location variations.

\begin{figure}[t]
    \centering
    \vspace{0.08in}
    \resizebox{\columnwidth}{!}{
        \begin{tabular}{@{}c@{\hspace{1mm}}c@{}}
            \includegraphics[width=0.4\columnwidth]{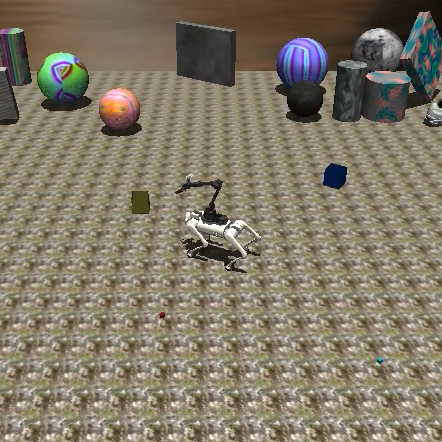}&
            \includegraphics[width=0.4\columnwidth]{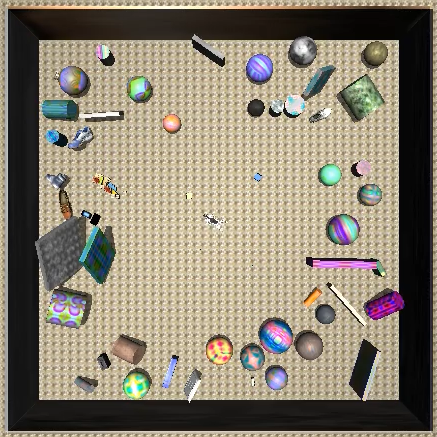}\\
        \end{tabular}
    }
    \vspace{-0.05in}
    \caption{\textbf{Example Training Scene with Randomized Objects.} Left: third-person view; Right: bird's-eye view.
    }
    \vspace{-0.25in}
    \label{fig:rand_background_objs}
\end{figure}

{\noindent \emph{C.2} \bf{Visual Gap.}} 
This gap is caused by a visual distribution mismatch between simulated and real-world pixels.
Since we do not assume knowing the target scenes in advance, we have to ensure that the perception model is able to handle a wide range of visual scenes.
For this, we use 1) a \emph{visual information bottleneck} (a pair of segmentation and depth maps) between the perception module and the policy, to reduce the visual \simtoreal{} gap while achieving better interpretability, 2) color modeling and randomization, by randomizing the colors seeded with real-world samples in the HSV space during simulation training, 
3)~visual augmentation including texture and background objects (\Figure~\ref{fig:rand_background_objs}) randomization as well as image pixel and spatial randomization~\cite{fan2021secant, drqv1}.

\begin{table*}[t]
\renewcommand{\arraystretch}{1.2}
\centering
\vspace{0.05in}
\caption{Real-world task success rates and episode time (mean$\pm$stddev over 3 random seeds), with 20 episodes per seed.}
\resizebox{\textwidth}{!}{
\begin{tabular}{l|c| D{,}{\pm}{-1} D{,}{\pm}{-1} D{,}{\pm}{-1} D{,}{\pm}{-1}|D{,}{\pm}{-1}}
\toprule
\multirow{2}{*}{\textbf{Method}} 
& \multirow{2}{*}{\textbf{Autonomous}}
& \multicolumn{4}{c|}{\textbf{Cumulative Success Rate} [\%] $\uparrow$} 
& \multicolumn{1}{c}{\multirow{2}{*}{\textbf{Episode Time} [s] $\downarrow$}} \\

&  & \multicolumn{1}{c}{\texttt{Search+MoveTo}} 
& \multicolumn{1}{c}{\texttt{Grasp}} 
& \multicolumn{1}{c}{\texttt{Search+MoveTo(WObj)}} 
& \multicolumn{1}{c|}{\texttt{DropInto} (Full Task)} 
& \\
\midrule
\textbf{\NoArmRetract} & \tikzcmark & \bkcolor{100.0},\bkcolor{0.0} & 58.3,51.1 & 28.3,40.7 & 5.0,8.7 & 87.3,4.8 \\
\textbf{\NoPerturb} & \tikzcmark & \bkcolor{100.0},\bkcolor{0.0} & 53.3,46.5 & 43.3,37.5 & 43.3,37.5 & 62.4,23.9 \\
\textbf{\NoVisualAug} & \tikzcmark  & 98.3,2.4 & 88.3,4.7 & 63.3,26.6 & 56.6,22.5 & 52.9,18.7 \\
\textbf{\DistillationOnly} & \tikzcmark & 93.3,4.7 & 71.1,33.2 & 60.0,38.9 & 50.0,35.5 & 61.2,25.5 \\
\cmidrule(lr){1-7}
\textbf{\HumanTeleop}  & \tikzxmark  & \bkcolor{100.0},\bkcolor{0.0} & \bkcolor{96.7},\bkcolor{2.9} & 86.7,7.6 & 75.0,5.0 & 65.5,3.6 \\
\cmidrule(lr){1-7}
\textbf{\name{} (ours)} & \tikzcmark & \bkcolor{100.0},\bkcolor{0.0} & \bkcolor{96.7},\bkcolor{5.8} & \bkcolor{96.7},\bkcolor{5.8} & \bkcolor{78.3},\bkcolor{5.8} & \bkcolor{43.8},\bkcolor{6.0} \\
\bottomrule
\end{tabular}
}
\vspace{-0.15in}
\label{tab:real_world_experimental_results}
\end{table*}

\vspace{-0.005in}
\section{Experiments and Results}
\vspace{-0.005in}
\label{sec:result}

{\textbf{Hardware and Sensors for Deployment}}: 
Our robotic system uses a Unitree Go1 with a top-mounted WidowX-250S manipulator. 
An Intel RealSense D435 camera is attached to the WidowX's wrist via a 3D-printed mount (\Figure~\ref{fig:robot_and_scene_config}).

{\textbf{Computational Resources for Training}}: 
All the models were trained with V100 or similar GPUs.
For teacher training, we used 4 GPUs, $120$ parallel environments, training iterations of $4$M ($\sim  60$M environment steps), and takes about 1 week. 
For student training, we used used 4 GPUs, 60 parallel environments, training iteration of $1.3$M ($\sim 40$M environment steps), and takes for about 2 weeks.
Low-level policy is trained with a single GPU, 5000 parallel environments and 4B environment steps for 3 days.

{\noindent \textbf{Computational Resources for Deployment}}: 
All model inference is performed on a laptop with a 12$^\textrm{th}$ Gen Intel i9-12900H CPU, NVIDIA RTX 3070Ti laptop GPU, and 16GB of RAM.
To minimize latency, we execute high- and low-level inference asynchronously at a frequency of 10 and 50\,Hz, respectively. 
The outputs from the models are then sent through the cable to the Raspberry PI  running our custom arm and quadruped drivers operating at approximately 500\,Hz. 


\noindent{\noindent \textbf{Baselines}}.
We compare \name{} against the following baselines. Some factors (\emph{e.g.} the visual bottleneck, stationary manipulation, rotational search) are essential for both safe and successful deployment with meaningful performances in real, and are applied to all the methods, thus do not appear in Table~\ref{tab:real_world_experimental_results}.
These ablation baselines are chosen due to their significant impact on top of the overall system with the common factors.

\vspace{-0.02in}
\begin{itemize}
    \item \NoArmRetract: Remove the arm retract (raise) reward in RL training.
    This baseline is included to show that in RL we might need to incorporate additional rewards for gearing the solutions towards a preferred solution space.
    \item \NoPerturb: Remove arm control randomization, arm mount perturbation, and object perturbation.
    \item \NoVisualAug: Remove random background objects and spatial visual augmentations.
    \item \DistillationOnly: Train student without the RL loss, keeping only the distillation and representation loss.
    \item \NoDistillation{} (end-to-end high-level training): Trains the student policy directly via RL without distillation from the teacher policy (\emph{i.e.,} RL for visuomotor policy learning from scratch). This baseline cannot learn to solve the task at all in simulation, because of the compounded difficulties of visual representation learning, behavior learning, and long-horizon exploration. It is excluded from the results.
    \item \HumanTeleop: An expert human teleoperator that provides locomotion commands and delta EE pose to the robot via a joystick. Delta EE is handled via IK. The teleoperator shared the same observation space as the policy, 
    \emph{i.e.,} observing solely through the wrist camera stream,
    and was allowed to practice on the system for an hour.
    While not fully autonomous, human teleoperation provides a valuable reference point to compare our model against.
\end{itemize}

\vspace{-0.15in}
\subsection{Results and Analysis}\label{sec:main_results}

To accurately evaluate policy performance, we run the entire training pipeline across three different random seeds for all methods, training from scratch the low-level, teacher, and student policies. 
Note that high-level policies of all different methods for one particular random seed share the same low-level policy trained using that seed.
We report the following two key metrics when deploying the final student  policy (together with the low-level policy used in training):
\begin{enumerate}
    \item \textbf{Cumulative Subtask Success Rate}:
    The success rate up to each subtask from the beginning of the episode.
    \item \textbf{Episode Time}: The time spent completing the full task from the beginning. We apply a time limit of $t_{\rm max}\!=\!90$\,s. For failed episodes, we use $t_{\rm max}$ as the episode time.
\end{enumerate}

\begin{figure}[t]
    \centering
    \vspace{0.08in}
    \begin{overpic}[width=8.5cm]{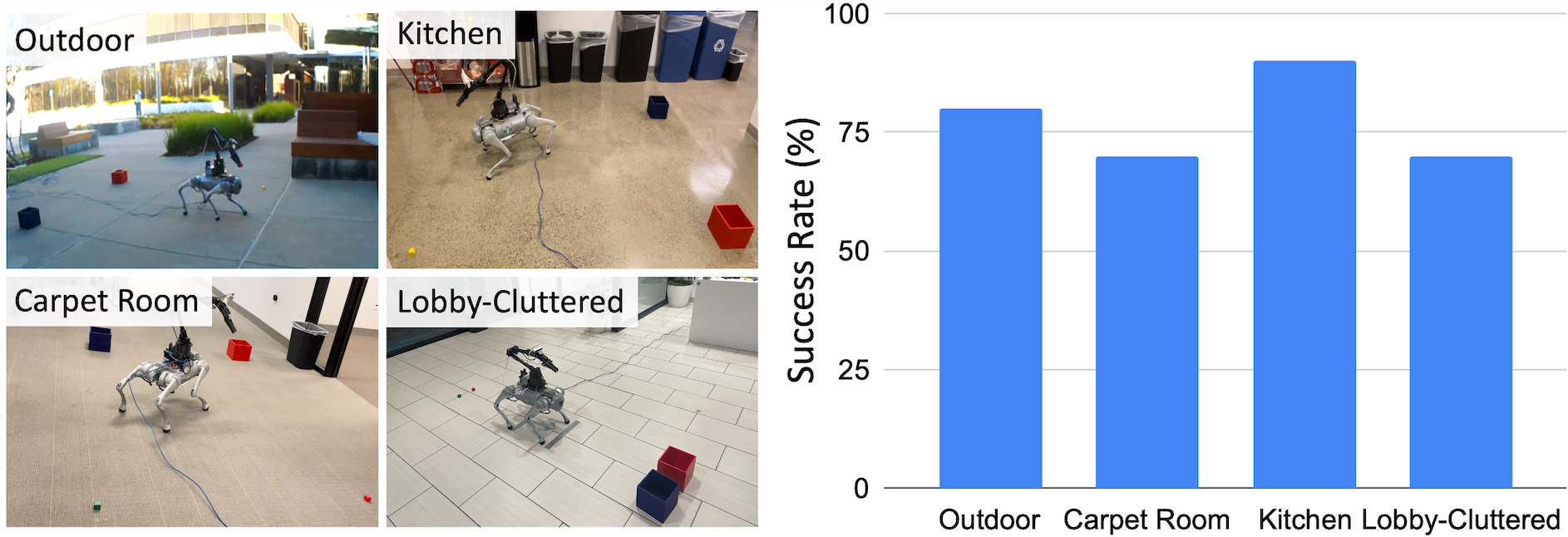}
    \end{overpic}
    \vspace{-0.05in}
    \caption{\textbf{The \emph{same} \name{} Policy across Scene Variations.} Left: images showing the differences for evaluation. Right: the average full task success rate across 10 real-world deployment episodes for each of the scenes shown on the left.}
    \vspace{-0.2in}
    \label{fig:scene_variation_results}
\end{figure}

{\flushleft For the main evaluation, we use the {\ttfamily Standard} object spatial layout in an indoor {\small \textsf{Lobby}} scene, as shown in \Figure~\ref{fig:robot_and_scene_config}, to maintain repeatability across all methods and seeds.}
For task objects, we use cubes and baskets of different colors.
For each method of each random seed, we roll out the policy for 20 episodes with varying object colors and positions according to a consistent evaluation protocol. This amounts to 360 real-world episodes in Table~\ref{tab:real_world_experimental_results}.
In addition to the {\ttfamily Standard} spatial setup in {\small \textsf{Lobby}}, we also evaluate the robot in several other scene and layout combinations in subsequent experiments, with another 40 real-world episodes (\Figure~\ref{fig:scene_variation_results}), leading to a total of 400 real-world episodes of experiments.

Table~\ref{tab:real_world_experimental_results} presents the main results, comparing \name{} against baselines under the {\ttfamily Standard} layout in the {\small \textsf{Lobby}} scene. 
As the ablations show, all the ablating factors  contribute to the efficient and successful completion of the final task. 

In particular, \NoArmRetract{} performs fairly well until \Grasp{}, when it only successfully locates 
the container and transports the cube to the container half of the times after a successful grasp.
There are several reasons for the failures. 
For one, without the arm retract reward acting as a soft constraint on the arm pose, the arm can remain at over-extended, close to to the ground.  This is very close to the physical torque limits of the shoulder motors responsible for raising the arm, leading to more arm shake, which in turn can result in the grasped cube slipping out of the fingers, or even motor failures. 
Secondly, right after grasp, the gripper and the camera are still relatively close to the ground, resulting in a low vantage view, and a higher chance of search failures.

Another baseline, \NoPerturb{}, has the lowest grasp success. 
This is because without perturbations of any kind (object, arm mount, or control), the policy tends to remember a fixed trajectory of grasping, which is not robust against the variations in the location of the object relative to the robot. In real deployments, the quadruped can stop a bit farther away from the cube, and the policy would move the gripper fingers toward the cube without actually reaching.
Lack of variation in object relative positions is likely because during simulation training, the policy tends to stop just at the success boundary of the \MoveTo{} subtask, so that the \Grasp{} subtask usually starts with very similar locations of the target object relative to the robot, while in the real world such clear cut positioning is rarely seen.

Success of \NoVisualAug{} suffers most during the \Search{} subtasks where the model sometimes gets distracted, due to false positive detections of the target in the background. 
There's a particularly large drop of roughly 25\% when searching for the basket.
This indicates that visual augmentations are important for the performance of the  vision module.

\DistillationOnly{} simply distills from the behavior of the highly successful teacher policy without using any task reward during student training. It can still achieve a reasonable performance, with a high variance across seeds.

In comparison, our full method \textbf{\name{}} uses distillation-guided RL, resulting in consistently high success across all seeds. 
It shows that distillation loss alone is not enough for learning a good policy. 
Task completion is also faster with the RL loss optimizing the policy further.

Overall, \textbf{\name{}} achieves the highest subtask and full task success rates among all the methods including \HumanTeleop, and is also the most efficient, taking on average 43.8 seconds to complete the full task, which is about 1.5$\times$ the speed of expert human teleoperation.

\vspace{-0.08in}
\subsection{Sim \emph{vs.} Real Performance}
\begin{figure}[h]
    \centering
    \vspace{-0.05in}
    \begin{overpic}[width=8cm]{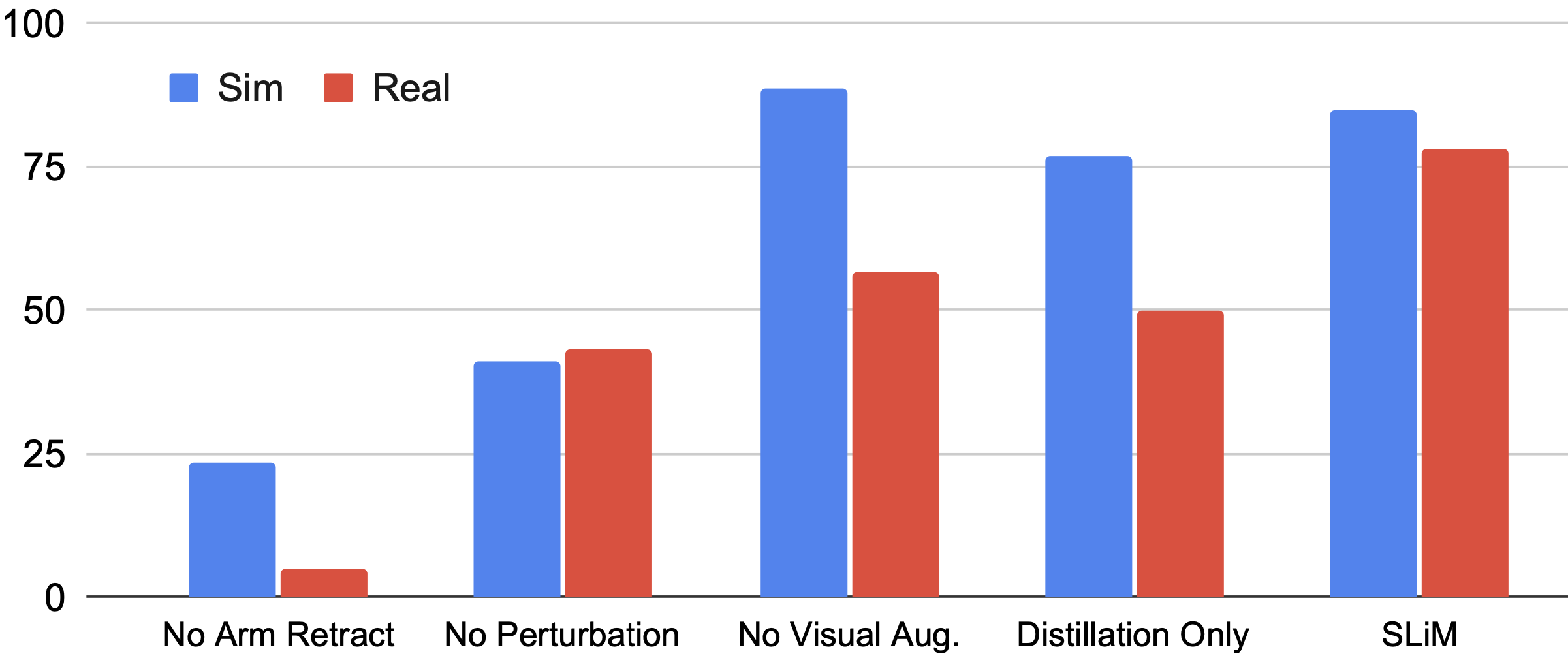}
    \end{overpic}
    \vspace{-0.05in}
    \caption{\textbf{Full Task Sim and Real Success Rate Comparison.}  The averaged success rate across three random seeds is used.}
    \vspace{-0.25in}
    \label{fig:sim_and_real_bar_plot_small_margin}
\end{figure}

\vspace{-0.05in}
To better understand how each technique impacts final success, we showcase the performance of all methods in both simulation and real in \Figure~\ref{fig:sim_and_real_bar_plot_small_margin}.
The sim success rate for each method denotes the average full task success rate over 100 episodes run in simulation, averaged across the three random seeds.  The simulation environment follows the same randomization setting as training, with the policy rollout being fully greedy.
The real success rate denotes the full task success rates from Table~\ref{tab:real_world_experimental_results}.

Without arm retract (\NoArmRetract{}), searching for the basket becomes more difficult in simulation as well as in real, because after grasping, the camera still points down to the ground, but searching for the basket requires many steps of arm raise. Without explicit rewards encouraging arm raise, it can create an exploration bottleneck for RL. 
Real-world performance suffers more, because moving with an out-stretched arm causes large sim-to-real gaps: the tip of the arm tends to shake more due to motor backlash, and the shoulder motors operate close to their physical torque limits, leading to more motor overloading.
For \NoPerturb{}, while one might expect perturbations to complicate RL learning, their inclusion results in much higher performance even in simulation. Without perturbations, the policy seems to converge to suboptimal solutions maybe due to a lack of exposure to a sufficiently large range of setups.
Without visual augmentations (\NoVisualAug{}), the simulation task is easier to be solved, and thus simulation success is artificially high, but real-world performance suffers when compared to the full method. RL training loss in the full \name{} method increases success in simulation (over \DistillationOnly{}) and achieves the highest success rate in real.

\subsection{Generalization to Different Real-world Scenes}
In addition to the standard scene used in Table~\ref{tab:real_world_experimental_results}, we further run real-world evaluations of \name{} 
under more scene variations (left side of \Figure~\ref{fig:scene_variation_results}): {\footnotesize\textsf{Outdoor}}, {\footnotesize\textsf{Carpet Room}}, {\footnotesize\textsf{Kitchen}}, and {\footnotesize\textsf{Lobby-Cluttered}}.
For the first three scenes, we use the same {\footnotesize \texttt{Standard}} spatial layout as before.
For the {\footnotesize\textsf{Lobby-Cluttered}} scene, we use the OOD {\footnotesize\texttt{Cluttered}} spatial layout (right side of \Figure~\ref{fig:robot_and_scene_config}).
We conduct ten trials for each environment using the same \name{} policy.

All results for these four additional scenes are summarized on the right side of \Figure~\ref{fig:scene_variation_results}.
For the three novel scenes, we observe that success rates are all fairly close to the results from Table~\ref{tab:real_world_experimental_results} ($\sim$78\%).
For the {\footnotesize \textsf{Lobby-Cluttered}} scene, the success rate is just a bit lower than that of the {\footnotesize\texttt{Standard}} layout due to being out of distribution.
This shows that the model trained fully in simulation can \emph{zero-shot} adapt to a wide range of real world settings, with vastly different lighting conditions, backgrounds, and floor types and textures.

\begin{figure}[h]
	\vspace{0.05in}
	\centering
	\begin{overpic}[width=8.5cm]{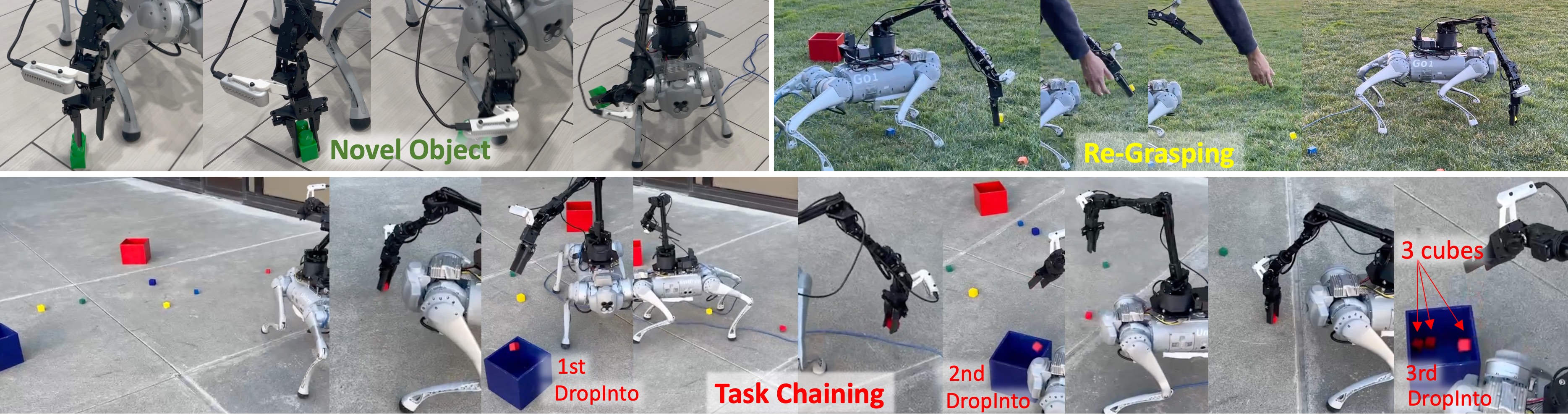}
	\end{overpic}
	\vspace{-0.03in}
	\caption{\textbf{Generalization Behaviors of \name{} Policy.}
		{\scriptsize \bf{\textsf{Novel Object}}}: grasping an object with a novel shape that is out of the training distribution.
		{\scriptsize \bf{\textsf{Re-Grasping}}}: a human interrupts the task progress by
		removing the cube from the gripper and tossing it to the ground. The \name{} policy will re-grasp the cube. 
		{\scriptsize \bf{\textsf{Task Chaining}}}: executing the full task multiple times (3 times in this example) consecutively without stopping. Each time a target cube (red cube) is dropped into the basket, another target cube is tossed onto the ground, and the \name{} policy will execute another full task starting from that state. 
	}
	\vspace{-0.2in}
	\label{fig:generalization}
\end{figure}

Furthermore, our qualitative evaluation showed that the robot also generalizes well to scenarios where:
\begin{enumerate}
\item task objects are randomly scattered on the ground (last two rows in \Figure~\ref{fig:multi_stage_pick_and_place}),
\item distractor objects are present (last row in \Figure~\ref{fig:multi_stage_pick_and_place}),
\item novel task object shapes (\Figure~\ref{fig:generalization}    {{\footnotesize \textsf{Novel Object}}}), 
\item a human interrupts the task progress (\Figure~\ref{fig:generalization} {\footnotesize \textsf{Re-Grasping}}),
\item and the robot executes the full task multiple times consecutively non-stop (\Figure~\ref{fig:generalization} {\footnotesize \textsf{Task Chaining}}).\end{enumerate}
Please refer to \href{https://horizonrobotics.github.io/gail/SLIM}{project page} for videos of emergent behaviors.

\vspace{-0.12in}
\subsection{Failure Modes}
\vspace{-0.03in}
We summarize and report the most common failure cases of \name{} here.  In one case, 
the quadruped stopped a bit late during grasping and kicked the cube away.  In another case, the low-level policy shook and moved backward, causing the cube to be just out of reach. \DropInto{} sometimes failed because of either early dropping right outside of the basket, or the quadruped kicking the basket away, right before dropping, when backing towards the basket.  In rare cases where the gripper held only the top part of the cube, the gripper fingers blocked the cube from the camera view, and the robot got stuck hovering over the basket without dropping the cube.

\section{Conclusion, Discussion and Future Work}

We present a legged mobile manipulation system for solving language-instructed long-horizon tasks.
The policy is trained via RL in sim with \simtoreal{} transfer, with a progressive policy expansion-based teacher policy for solving the long-horizon task, followed by a distillation-guided RL trained student visuomotor policy. We further identify and design a suite of crucial techniques to reduce the \simtoreal{} gap.
Experimental results against various baseline methods verify the effectiveness of the proposed \name{} system.
Moreover, real-world testing of \name{} across different scenes and spatial layouts
shows that the robot can generalize well across scenes.

For future work, we plan to further expand the applicability of \name{} across more diverse scenarios by improving locomotion policy training, increasing visual and language diversity, and extending to a wider range of objects and tasks.


\vspace{-0.05in}
\bibliographystyle{IEEEtran}
\bibliography{ref}

\begin{thebibliography}{10}
\providecommand{\url}[1]{#1}
\csname url@rmstyle\endcsname
\providecommand{\newblock}{\relax}
\providecommand{\bibinfo}[2]{#2}
\providecommand\BIBentrySTDinterwordspacing{\spaceskip=0pt\relax}
\providecommand\BIBentryALTinterwordstretchfactor{4}
\providecommand\BIBentryALTinterwordspacing{\spaceskip=\fontdimen2\font plus
\BIBentryALTinterwordstretchfactor\fontdimen3\font minus
  \fontdimen4\font\relax}
\providecommand\BIBforeignlanguage[2]{{%
\expandafter\ifx\csname l@#1\endcsname\relax
\typeout{** WARNING: IEEEtran.bst: No hyphenation pattern has been}%
\typeout{** loaded for the language `#1'. Using the pattern for}%
\typeout{** the default language instead.}%
\else
\language=\csname l@#1\endcsname
\fi
#2}}

\bibitem{deep_wholebody}
Z.~Fu, X.~Cheng, and D.~Pathak, ``Deep whole-body control: Learning a unified
  policy for manipulation and locomotion,'' in \emph{Conference on Robot
  Learning}, 2022.

\bibitem{ASC}
N.~Yokoyama, A.~W. Clegg, J.~Truong, E.~Undersander, J.~Yang, S.~Arnaud, S.~Ha,
  D.~Batra, and A.~Rai, ``{{ASC}: Adaptive Skill Coordination for Robotic
  Mobile Manipulation},'' \emph{IEEE Robotics and Automation Letters}, vol.~9,
  no.~1, pp. 779--786, 2024.

\bibitem{pan2024roboduet}
G.~Pan, Q.~Ben, Z.~Yuan, G.~Jiang, Y.~Ji, J.~Pang, H.~Liu, and H.~Xu,
  ``{RoboDuet}: Learning a cooperative policy for whole-body legged
  loco-manipulation,'' \emph{IEEE Robotics and Automation Letters}, vol.~10,
  no.~5, pp. 4564--4571, 2025.

\bibitem{VBC}
M.~Liu, Z.~Chen, X.~Cheng, Y.~Ji, R.~Yang, and X.~Wang, ``Visual whole-body
  control for legged loco-manipulation,'' in \emph{Conference on Robot
  Learning}, 2024.

\bibitem{GEFF}
R.-Z. Qiu, Y.~Hu, G.~Yang, Y.~Song, Y.~Fu, J.~Ye, J.~Mu, R.~Yang, N.~Atanasov,
  S.~Scherer, and X.~Wang, ``Learning generalizable feature fields for mobile
  manipulation,'' \emph{CoRR}, vol. arXiv:2403.07563, 2024.

\bibitem{GAMMA}
J.~Zhang, N.~Gireesh, J.~Wang, X.~Fang, C.~Xu, and W.~Chen, ``{GAMMA}:
  Graspability-aware mobile manipulation policy learning based on online
  grasping pose fusion,'' in \emph{IEEE International Conference on Robotics
  and Automation}, 2024.

\bibitem{review1}
Y.~Gong, G.~Sun, A.~Nair, A.~Bidwai, R.~CS, J.~Grezmak, G.~Sartoretti, and
  K.~A. Daltorio, ``Legged robots for object manipulation: A review,''
  \emph{Frontiers in Mechanical Engineering}, vol. Volume 9 - 2023, 2023.

\bibitem{review2}
S.~S.~K. et~al., ``Next generation legged robot locomotion: A review on control
  techniques,'' \emph{Heliyon}, vol.~10, no.~18, p. e37237, 2024.

\bibitem{sleiman2023wholebody_control}
J.-P. Sleiman, F.~Farshidian, and M.~Hutter, ``Versatile multicontact planning
  and control for legged loco-manipulation,'' \emph{Science Robotics}, vol.~8,
  no.~81, p. eadg5014, 2023.

\bibitem{ma2022mpc}
Y.~Ma, F.~Farshidian, T.~Miki, J.~Lee, and M.~Hutter, ``Combining
  learning-based locomotion policy with model-based manipulation for legged
  mobile manipulators,'' \emph{IEEE Robotics and Automation Letters}, vol.~7,
  no.~2, pp. 2377--2384, 2022.

\bibitem{huang2024manipulatortail}
H.~Huang, A.~Loquercio, A.~Kumar, N.~Thakkar, K.~Goldberg, and J.~Malik,
  ``Manipulator as a tail: Promoting dynamic stability for legged locomotion,''
  in \emph{IEEE International Conference on Robotics and Automation}, 2024.

\bibitem{wholebody}
S.~Jeon, M.~Jung, S.~Choi, B.~Kim, and J.~Hwangbo, ``Learning whole-body
  manipulation for quadrupedal robot,'' \emph{IEEE Robotics and Automation
  Letters}, vol.~9, no.~1, pp. 699--706, 2024.

\bibitem{zhang2024learning}
M.~Zhang, Y.~Ma, T.~Miki, and M.~Hutter, ``Learning to open and traverse doors
  with a legged manipulator,'' in \emph{Conference on Robot Learning}, 2024.

\bibitem{wu2024helpfuldoggybot}
Q.~Wu, Z.~Fu, X.~Cheng, X.~Wang, and C.~Finn, ``{Helpful} {DoggyBot}:
  Open-world object fetching using legged robots and vision-language models,''
  \emph{CoRR}, vol. arXiv:2410.00231, 2024.

\bibitem{tabletop_pick_place}
A.~S. Morgan, K.~Hang, W.~G. Bircher, F.~M. Alladkani, A.~Gandhi, B.~Calli, and
  A.~M. Dollar, ``Benchmarking cluttered robot pick-and-place manipulation with
  the box and blocks test,'' \emph{IEEE Robotics and Automation Letters},
  vol.~5, no.~2, pp. 454--461, 2020.

\bibitem{walk_these_ways}
G.~B. Margolis and P.~Agrawal, ``Walk these ways: Tuning robot control for
  generalization with multiplicity of behavior,'' in \emph{Conference on Robot
  Learning}, 2022.

\bibitem{fan2021secant}
L.~Fan, G.~Wang, D.-A. Huang, Z.~Yu, L.~Fei-Fei, Y.~Zhu, and A.~Anandkumar,
  ``Secant: Self-expert cloning for zero-shot generalization of visual
  policies,'' in \emph{International Conference on Machine Learning}, 2021.

\bibitem{capacity_loss}
C.~Lyle, M.~Rowland, and W.~Dabney, ``Understanding and preventing capacity
  loss in reinforcement learning,'' in \emph{International Conference on
  Learning Representations}, 2022.

\bibitem{understanding_plasticity}
C.~Lyle, Z.~Zheng, E.~Nikishin, B.~Avila~Pires, R.~Pascanu, and W.~Dabney,
  ``Understanding plasticity in neural networks,'' in \emph{International
  Conference on Machine Learning}, 2023.

\bibitem{pex}
H.~Zhang, W.~Xu, and H.~Yu, ``Policy expansion for bridging offline-to-online
  reinforcement learning,'' in \emph{International Conference on Learning
  Representations}, 2023.

\bibitem{sac}
T.~Haarnoja, A.~Zhou, P.~Abbeel, and S.~Levine, ``Soft actor-critic: Off-policy
  maximum entropy deep reinforcement learning with a stochastic actor,'' in
  \emph{International Conference on Machine Learning}, 2018.

\bibitem{metaworld}
T.~Yu, D.~Quillen, Z.~He, R.~Julian, K.~Hausman, C.~Finn, and S.~Levine,
  ``Meta-world: A benchmark and evaluation for multi-task and meta
  reinforcement learning,'' in \emph{Conference on Robot Learning}, 2019.

\bibitem{PaCo}
L.~Sun, H.~Zhang, W.~Xu, and M.~Tomizuka, ``{PaCo}: Parameter-compositional
  multi-task reinforcement learning,'' in \emph{Advances in Neural Information
  Processing Systems}, 2022.

\bibitem{yu2022you}
H.~Yu, H.~Zhang, and W.~Xu, ``Do you need the entropy reward (in practice)?''
  \emph{CoRR}, vol. arXiv:2201.12434, 2022.

\bibitem{drqv1}
D.~Yarats, I.~Kostrikov, and R.~Fergus, ``Image augmentation is all you need:
  Regularizing deep reinforcement learning from pixels,'' in
  \emph{International Conference on Learning Representations}, 2021.

\end{thebibliography}

\end{document}